\documentclass[twoside,11pt]{article}

\usepackage{jmlr2e}
\usepackage{times}
\usepackage{amsmath}
\usepackage{floatrow}
\usepackage{floatrow}
\usepackage{enumitem}
\usepackage{threeparttable}
\usepackage{booktabs}
\usepackage{bm}
\usepackage{wrapfig}
\usepackage[dvipsnames,svgnames,x11names]{xcolor}
\usepackage{soul}
\usepackage{subcaption}
\usepackage{placeins}
\usepackage{array}

\setlist[itemize]{noitemsep, nolistsep, leftmargin=*}
\setlist[enumerate]{noitemsep, nolistsep, leftmargin=*}

\begin{document}

\title{MetNet: A Neural Weather Model for Precipitation Forecasting}

\author{\name Casper Kaae S{\o}nderby\thanks{These authors contributed equally. Contributions are as follows. Project start: CKS. Project management: CKS, NK, AO, LE, TS, JaH. Data and infrastructure: LE, CKS, JoH, MD, AO, SA, JaH. Model: NK, CKS, JoH, LE. Experiments: NK, CKS, JoH, LE, MD, TS. Paper: NK, CKS and LE. Contact: casperkaae@google.com}
\AND \name Lasse Espeholt \AND \name  Jonathan Heek \AND \name Mostafa Dehghani  \AND \name Avital Oliver \hfill \textnormal{\textsc{Google Research}}  \AND \name Tim Salimans \AND \name Shreya Agrawal \AND \name Jason Hickey \AND \name Nal Kalchbrenner$^*$  }

\maketitle

\begin{abstract}
Weather forecasting is a long standing scientific challenge with direct social and economic impact. The task is suitable for deep neural networks due to vast amounts of continuously collected data and a rich spatial and temporal structure that presents long range dependencies. We introduce MetNet, a neural network that forecasts precipitation up to 8 hours into the future at the high spatial resolution of 1 km$^2$ and at the temporal resolution of 2 minutes with a latency in the order of seconds. MetNet takes as input radar and satellite data and forecast lead time and produces a probabilistic precipitation map. The architecture uses axial self-attention to aggregate the global context from a large input patch corresponding to a million square kilometers. We evaluate the performance of MetNet at various precipitation thresholds and find that MetNet outperforms Numerical Weather Prediction at forecasts of up to 7 to 8 hours on the scale of the continental United States. 
\end{abstract}

\section{Introduction}

Weather forecasting is the prediction of future weather conditions such as precipitation, temperature, pressure and wind and is fundamental to both science and society. The field joins the forefronts of physical modelling and computing technology into a single century-long scientific and technological endeavor \citep{bauer2015quiet}. The social and economic benefits of accurate weather forecasting range from improvements in our daily lives to substantial impacts on agriculture, energy and transportation and to the prevention of human and economic losses through better prediction of hazardous conditions such as storms and floods~\citep{shakti2015comparison, hwang2015improved}.

Operational weather forecasts are based on Numerical Weather Prediction (NWP) using the laws of physics to simulate the dynamics of the atmosphere. NWP has seen substantial advances over the preceding decades due to improvements in the representation of the physics, an increase in observational data and an exponential growth in computing capabilities. These aspects have increased the spatial and temporal resolution of NWP and have advanced the range of skillful weather forecasts by about one day per decade. Despite the advances, several challenges remain for NWP~\citep{bauer2015quiet}. The computational and power demands of NWP grow as a power of the resolution of the forecast and create a trade-off between the accuracy of the forecast that requires increasing levels of resolution and the time required to make the forecast. The mathematical formulation of NWP is derived from our current understanding of atmospheric physics, which might be imprecise or cannot be fully resolved at the resolution of the model.

\begin{figure*}[t]
    \begin{subfigure}[b]{.49\textwidth}
        \centering
        \includegraphics[width=1\textwidth, height=.4\textwidth]{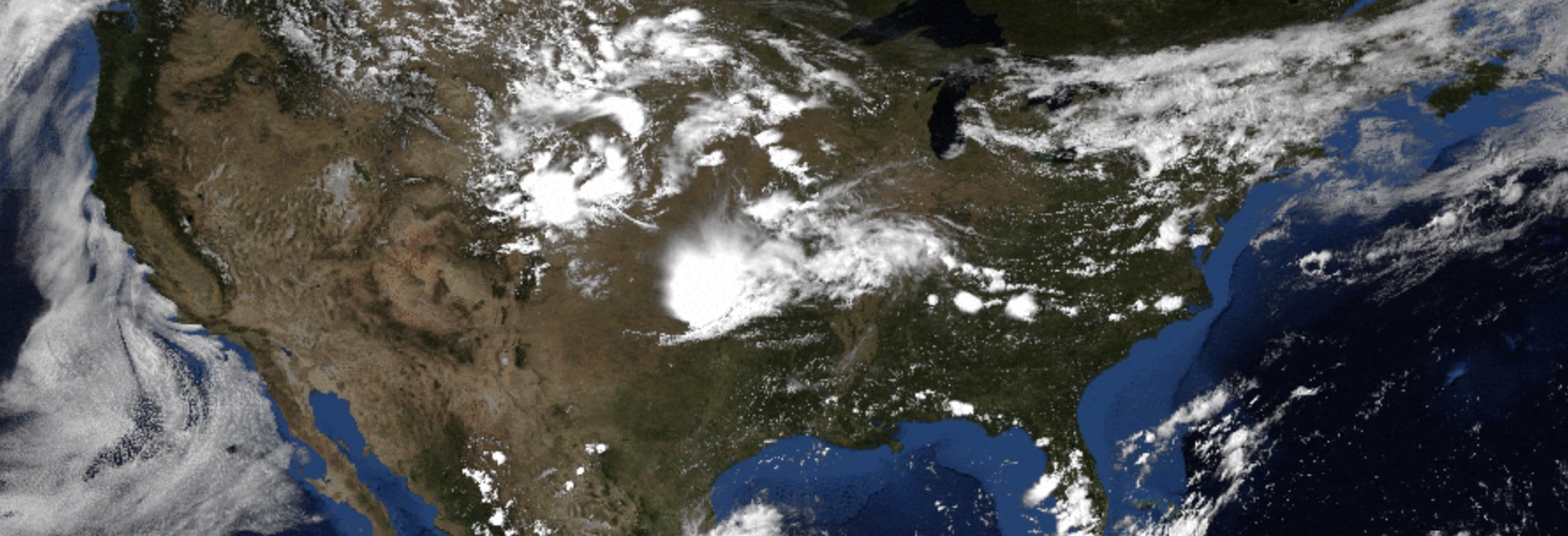}
    \end{subfigure}\,\,\,%
    \begin{subfigure}[b]{.49\textwidth}
        \centering
        \includegraphics[width=1\textwidth, height=.4\textwidth]{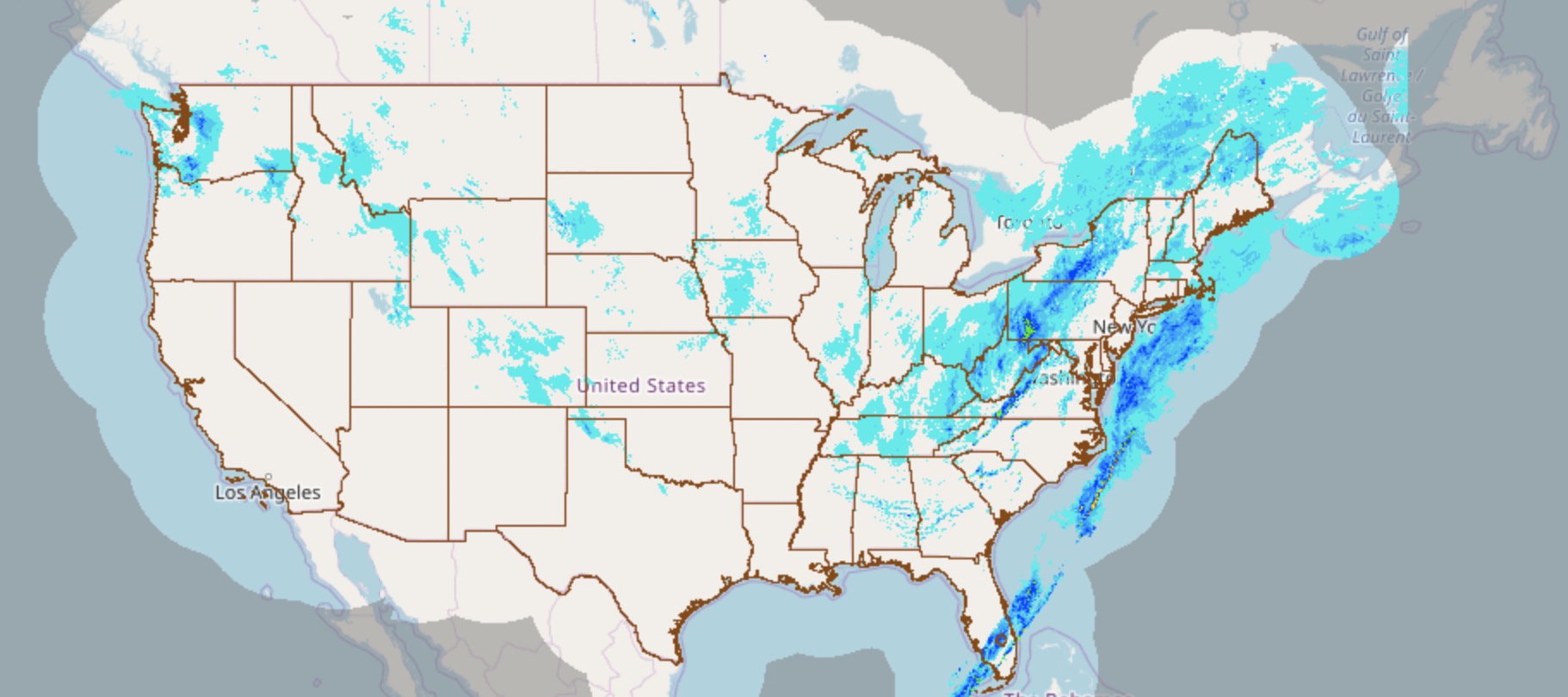}
        
    \end{subfigure}%
    \caption{Data sources. Left: GOES-16 visual bands. Right: MRMS precipitation rates. (Preliminary)}
    \label{fig:conus}
\end{figure*}

In the field of machine learning deep neural networks (DNN) have seen remarkable progress in recent years due to increased amounts of available data, better model architectures and ease of implementation on powerful specialized hardware such as GPUs and TPUs~\citep{jouppi2017datacenter}.
Recently introduced DNN architectures are able to effectively process and make use of large spatial and temporal contexts in the input data~\citep{krizhevsky2012imagenet,he2016deep,vaswani2017attention}. These models can produce probabilistic outputs and represent uncertainty in the predictions. The predictive performance tends to improve with increasing amounts of training data while the specification of the DNNs remains easy to comprehend and maintain. These properties make DNNs especially promising for weather forecasting due to the vast amounts of continually collected data from satellites, ground based radars and weather stations requiring no human annotation,  the large computational requirements of the task,  the rich spatial and temporal structure of the inputs and predictions and  the inherent need to represent uncertainty.

We adopt DNNs to tackle the self-annotated structured prediction task of precipitation forecasting. The setting is an ideal benchmark due to the availability of dense and continual precipitation measurements that do not require human annotation and well defined metrics to measure performance. We develop \emph{MetNet} that is a Neural Weather Model (NWM) that forecasts rates of precipitation with a lead time of up to 8 hours, a spatial resolution of $1\,\mathrm{km}^2$ and a temporal resolution of $2$ minutes. MetNet covers a $7000 \times 2500$ km geographical area corresponding to the continental United States. The architecture is conditioned on lead time and uses axial self-attention \citep{ho2019axial} at its core to aggregate a large spatial context of $1024 \times 1024$ km. MetNet relies on mosaicked ground based radar and satellite imagery as input and the predictions take in the order of seconds independently of lead time and can be done in parallel. \

We show that for up to 7 to 8 hours of lead time MetNet is able to outperform the High Resolution Rapid Refresh (HRRR) system which is the current best operational NWP available from NOAA, the National Oceanic and Atmospheric Administration. MetNet substantially outperforms other 2-3 hour forecasting methods~\citep{ayzel2019optical, agrawalmachine}. Ablation studies show that MetNet's architecture is able to capture the large spatial context that is needed for accurate predictions with many hours of lead time. These results also suggest the effectiveness of axial self-attention blocks beyond generative modelling of images.  Visualizations suggest that MetNet is able to capture advection  and the formation of new regions of precipitation. To our knowledge MetNet is the first machine learning model to outperform NWP on a structured prediction task at such a range and resolution.

\section{Precipitation Forecasting}

\begin{figure*}
    \begin{subfigure}[b]{1.\textwidth}
        \centering
        \includegraphics[width=1\textwidth]{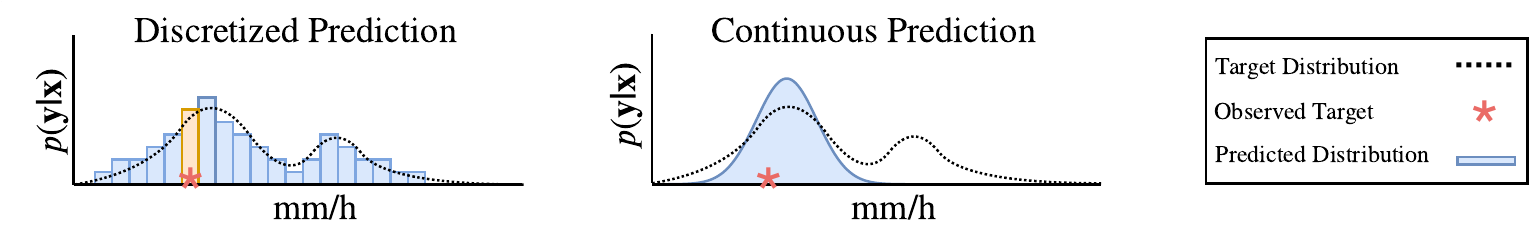}
    \end{subfigure}%

    \caption{Comparison of continuous and discrete predictions. The discrete prediction is arbitrarily flexible and better able to approximate a complex target distribution.\label{fig:distribution2}}
\end{figure*}

Precipitation provides a benchmark for a highly varying and densely measured target~\citep{agrawalmachine}. We cast precipitation forecasting as a structured prediction problem where the output comes in the form of a three-dimensional tensor. Each value of the tensor corresponds to a time and a location and indicates the corresponding rate of precipitation measured in mm/h. Target precipitation rates are estimated by the Multi Radar Multi Sensor (MRMS) ground based radars as a function of the returned radar echoes~\citep{zhang2016multi}. The spatial size obtained from MRMS is $7000 \times 2500$ covering the continental United States. Each pixel covers $0.01^\circ$ of longitude and latitude corresponding to approximately 1 km$^2$. In addition to MRMS frames, the available input data include the 16 spectral bands of the optical Geostationary Operational Environmental Satellite 16 (GOES-16). Figure 1 contains examples of MRMS and GOES-16 frames.

\section{Neural Weather Models}
We next describe defining features of DNNs when used as neural weather models (NWM) that is as models for the prediction of structured weather variables. 

\begin{figure}[t]
\includegraphics[width=1.0\textwidth]{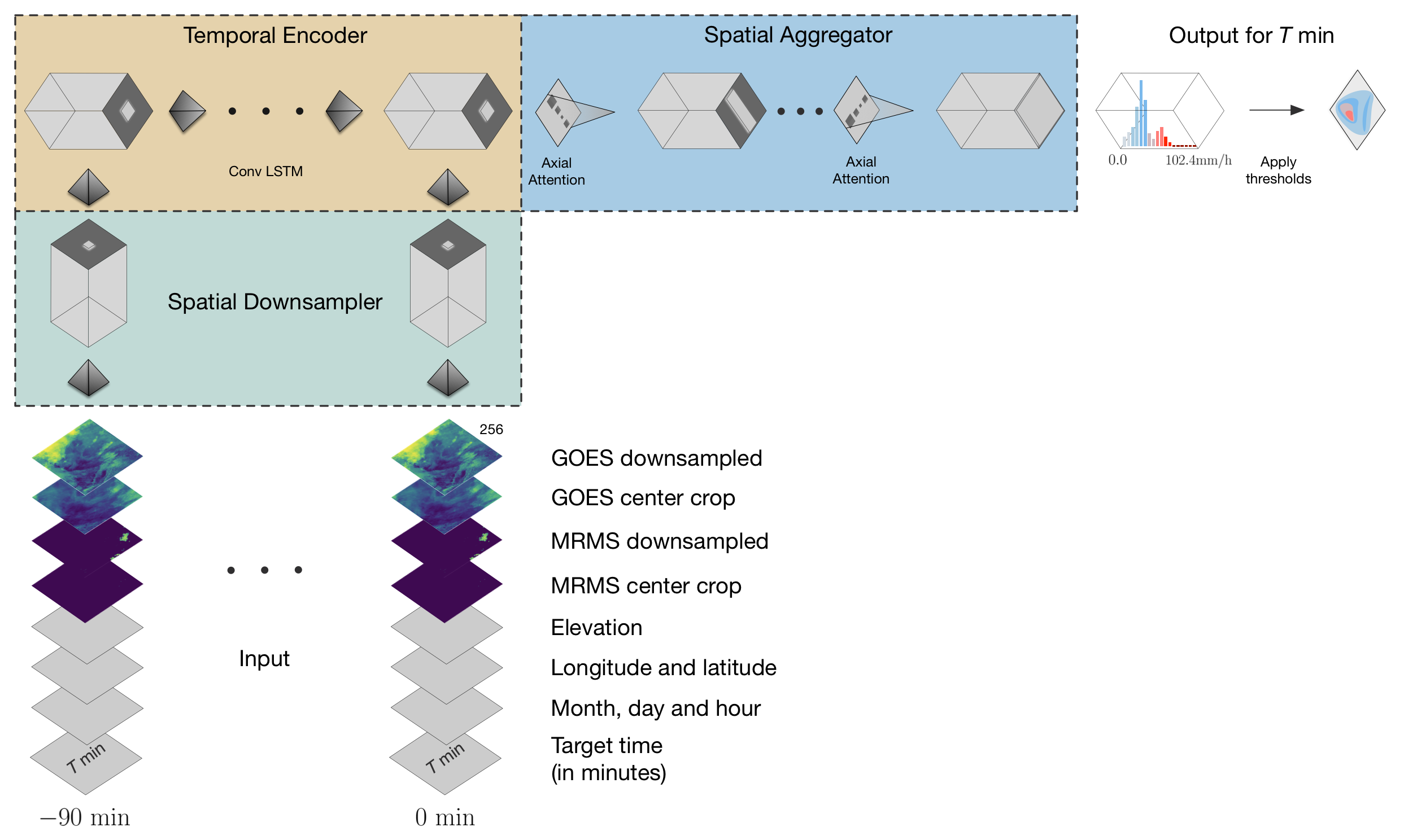}
\caption{\label{fig:architecture} Diagram of the input patch and the three parts of the MetNet architecture. The target lead time represented by the integer ${\bm{y}}$ is concatenated at the input to indicate to MetNet the desired lead time for the prediction. The Spatial Downsampler encodes the time slices sampled every 15 minutes using a shared convolutional neural network. The Temporal Encoder processes the downsampled time slices in the direction of time using a Convolutional LSTM. The Spatial Aggregator uses 8 axial self-attention blocks (only 2 shown) to cover a global receptive field over the input patch. 
}
\end{figure}

\subsection{Direct Probabilistic Model}
NWMs predict the probabilities of target weather conditions $\bm{y}$ at a target time $T_{\bm{y}}$ from a set of input conditions $\bm{x}$ at an input time up to $T_{\bm{x}}$:
\begin{align}
    p(\bm{y}|\bm{x}) = \mathrm{DNN}_{\bm{\theta}}(\bm{x}), \label{eq:prediction1}
\end{align}
where $p(\bm{y}|\bm{x})$ is a probability distribution over the targets $\bm{y}$ given the inputs $\bm{x}$ and $\mathrm{DNN}_{\bm{\theta}}$ is a deep neural network with learnable parameters $\theta$. The probabilistic form of a NWM accounts for uncertainty  by computing a probability distribution over possible outcomes. The NWM does not produce a single deterministic output.

\subsection{Architectural Constraints and Learning}
NWMs contain no explicit assumptions about the physics of weather. One encodes generic assumptions about the spatial and temporal relations among the inputs and between inputs and targets in the form of parametric architectural constraints in the NWM. One learns the parameters $\bm{\theta}$ governing these relations through back-propagation by minimizing the forecast error between the observed $\bm{y}$ and the predicted $p(\bm{y}|\bm{x})$.

\subsection{Discrete Distributions}
Weather variables often correspond to physical measurements on a continuous scale. Instead of modelling the continuous values of the targets $\bm{y}$ we discretize the variables into a large number of small intervals that cover the continuous range of interest. The prediction from the NWM is then a categorical distribution that assigns a probability to each of the intervals (Figure~\ref{fig:distribution2}). The generated categorical distribution is highly flexible and stabilises the training of the underlying DNN \citep{oord2016pixel, kalchbrenner2017video}.

\section{MetNet Architecture}

We next describe the neural network architecture that underlies the NWM for precipitation rate forecasting that we call MetNet.

\subsection{Input Patch}
\label{metnet:data}

MetNet receives as input a four-dimensional tensor of size $[t,w,h,c]$ that corresponds to data from a large patch of the continental United States with dimensions time, height, width and number of channels. 
The time dimension comprises $t$ slices sampled every 15 minutes over a 90 minutes interval prior to $T_{\bm{x}}$, where $T_{\bm{x}}$ is the time at which the model makes a prediction into the future. 
The input data is calculated from a patch covering a geographical area of $1024\times1024$ kilometers corresponding to $1024\times1024$ values. The input features comprise the single MRMS radar image, the 16 spectral bands of the GOES-16 satellite and additional real-valued features for the longitude, latitude and elevation of each location in the patch as well as for the hour, day and month of the input time $T_{\bm{x}}$. The latter time features are tiled along the spatial dimensions of the input tensor (Figure~\ref{fig:architecture}).

\subsection{Conditioning on Target Lead Time}

A forward pass through MetNet makes a prediction for a single lead time. The model is informed about the desired lead time by concatenating this information with the descriptive input features.
The aim is for the computation in MetNet to be aware of the lead time of the prediction from the very outset so that every aspect of the computation can be conditioned on the lead time. The lead time is represented as an integer $i = (T_{\bm{y}} / 2)-1$ indicating minutes from 2 to 480. The integer $i$ is tiled along the $w \times h$ locations in the patch and is represented as an all-zero vector with a 1 at position $i$ in the vector. By changing the target lead time given as input, one can use the same MetNet model to make forecasts for the entire range of target times that MetNet is trained on.
\subsection{Target Patch}

Accurate predictions of the target tensor with up to 480 minutes of lead time require a large spatial context around the target. With an input patch covering $1024 \times 1024$ km and an indicative average precipitation displacement of 1 km per minute, we set the target patch to cover $64\times64$ km centered on the input patch. This leaves at least 480 km of spatial context on each of the four sides of the target patch satisfying the indicative displacement rate.

\begin{figure*}
    \begin{subfigure}[b]{.35\textwidth}
        \centering
        \includegraphics[width=1\textwidth]{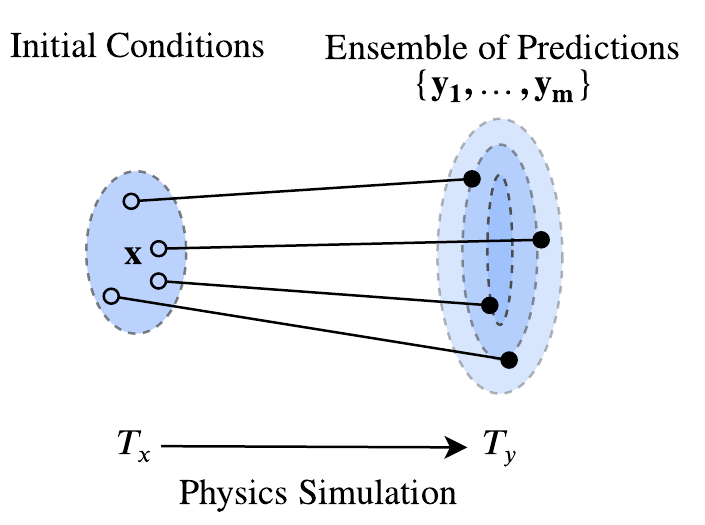}
    \end{subfigure}\hspace{1cm}%
    \begin{subfigure}[b]{.35\textwidth}
        \centering
        \includegraphics[width=1\textwidth]{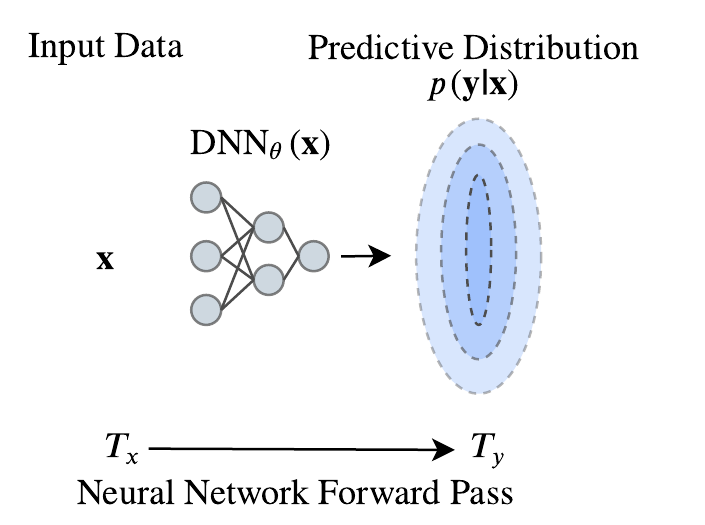}
    \end{subfigure}%
    \caption{Properties of NWP  and NWMs. Left: NWP performs a deterministic physical simulation starting from the initial conditions. The predictive uncertainty is estimated from an ensemble of predictions each run with slightly different initial conditions. Right: The NWM treats the current observations as direct inputs to a DNN, directly estimating the distribution over future conditions $p(\bm{y}|\bm{x})$. \label{fig:modelcomp2}}
\end{figure*}

\subsection{Output Layer}
The output of MetNet is a 512-way categorical distribution. Each of the 512 bins corresponds to a $0.2$ mm/h interval of predicted precipitation rate starting from $0$ mm/h to $102.4$ mm/h. All precipitation rates higher than $102.4$ mm/h are grouped in the last bin. Probabilities for any precipitation range of interest or for precipitation rate above a given threshold are obtained by summing probabilities of the intervals in that range or above that threshold.

\subsection{Spatial Downsampler}
\label{metnet:downsampler}

MetNet aims at fully capturing the spatial context in the input patch. A trade-off arises between the fidelity of the representation and the memory and computation required to compute it. To maintain viable memory and computation requirements, the first part of MetNet contracts the input tensor spatially using a series of convolution and pooling layers. The $t$ slices along the time dimension of the input patch are processed separately. Each slice is first packaged into an input tensor of spatial dimensions $256 \times 256$ (see Appendix~\ref{app:data} for the exact pre-processing operations). 
Each slice is then processed by the following neural network layers: a $3\times3$ convolution with 160 channels, a $2\times2$ max-pooling layer with stride 2, three more $3\times3$ convolutions with 256 channels and one more $2\times2$ max pooling layer with stride 2. These operations produce $t$ tensors of spatial dimensions $64 
\times 64$ and 256 channels.

\subsection{Temporal Encoder}
\label{metnet:convlstm}
The second part of MetNet encodes the input patch along the temporal dimension. The $t$ spatially contracted slices are given to  a recurrent neural network following the order of time. We use a  Convolutional Long Short-Term Memory network with kernel size $3 \times 3$ and 384 channels for the temporal encoding ~\citep{xingjian2015convolutional}. The recurrent  network is able to gauge the temporal dynamics of the input slices and by following the direction of time it is able to give more relevance to the patterns in the most recent input slice. The result is a single tensor of size $64 
\times 64$ and 384 channels, where each location summarizes spatially and temporally one region of the large context in the input patch.

\begin{figure}[t]
\includegraphics[width=1.0\textwidth]{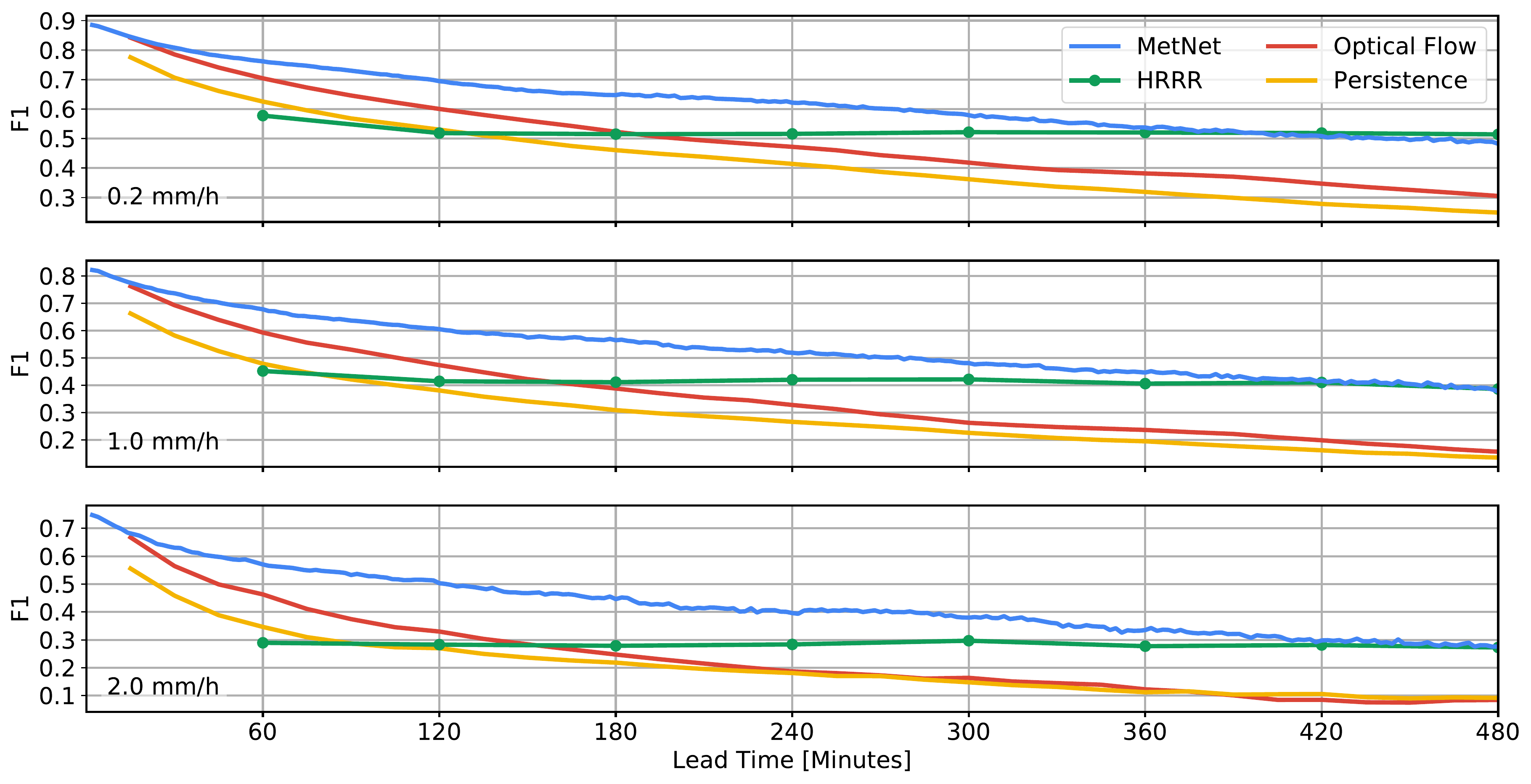}
\caption{ F1 scores for MetNet, HRRR, Optical Flow and Persistence for 2 to 480 minutes of lead time evaluated at $0.2$ mm/h (top), $1.0$ mm/h (middle) and $2.0$ mm/h (bottom) precipitation rates. MetNet outperforms HRRR up to 400 to 480 minutes and outperforms a strong optical flow method and the persistence baseline throughout the 480 minute range.
\label{fig:F1results}}
\end{figure} 

\subsection{Spatial Aggregator}
\label{metnet:attention}

To make MetNet's receptive field cover the full global spatial context in the input patch, the third part of MetNet uses a series of eight axial self-attention blocks ~\citep{ho2019axial, donahue2019large}. Four axial self-attention blocks operating along the width and four blocks operating along the height are interleaved and have 2048 channels and 16 attention heads each. Axial self-attention blocks sidestep the computationally prohibitive quadratic factor of vanilla self-attention blocks~\citep{vaswani2017attention} while preserving the benefit of reaching a full receptive field. The global context can be reached in just two axial self-attention blocks compared to the 32 blocks that are needed using standard $3\times3$ convolutions. 
In total this setting for MetNet has 225M parameters.

\begin{figure}[t]
\includegraphics[width=1.0\textwidth]{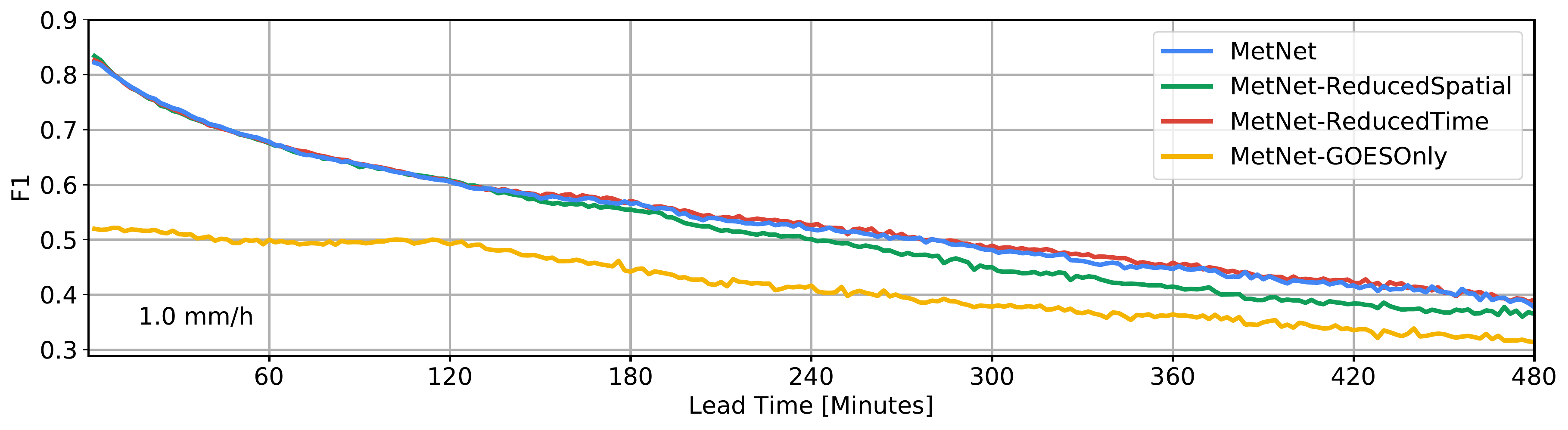}
\caption{\label{fig:ablations} F1-scores of the ablation experiments at the 1.0 mm/h precipitation rate threshold. MetNet-ReducedSpatial indicates the importance of capturing spatial context beyond 512 km in the input patch and in the architecture.}
\end{figure}

\begin{table}[]
\small
\begin{tabular}{lcclc}
\toprule
                     \textbf{Configuration}  & \textbf{Spatial Context (km)} & \textbf{Temporal Context (min)} & \textbf{Data Sources}       \\
                       \midrule
MetNet                 & 1024            & 90        & MRMS, GOES-16       \\
MetNet-ReducedSpatial  & 512             & 90        & MRMS, GOES-16       \\
MetNet-ReducedTemporal & 1024            & 30        & MRMS, GOES-16       \\
MetNet-GOESOnly        & 1024            & 90        & GOES-16             \\
\bottomrule
\end{tabular}
\caption{\label{tab:ablations} Specification of the various configurations used in the ablation experiments. The geographical features of longitude, latitude and elevation and the timestamp features of month, hour and day are included in all configurations.}
\end{table}

\section{Discussion of NWMs and NWP}

We next discuss some of the properties of MetNet and NWMs in relation to analogous aspects of NWP~\citep{bauer2015quiet}. MetNet outputs directly a probability distribution, while NWP produces a probabilistic prediction by ensembling a set of deterministic physical simulations with either different initial conditions  (Figure~\ref{fig:distribution2}) or model parameters. MetNet is constructed from general modules such as convolutions, recurrence and attention that depend on the structure of the task but are largely independent of the underlying domain, whereas NWP relies on explicit phenomena-dependent physical equations; a consequence of this is that MetNet's performance correlates with the availability of data for the task. The latency of MetNet is independent of the magnitude of the target lead time and all lead times can be predicted at once in parallel. On the other hand, in NWP the simulation is sequential along time and the latency scales linearly with target lead time. In practice MetNet's latency is in the order of seconds for any of the target lead times (up to 480 minutes), whereas that of NWP is in order of tens of minutes to hours. An interesting observation concerns the scalability of models as currently designed with respect to the underlying resolution. For MetNet a doubling in spatial resolution requires 4 times more computation using the current architectural choices. For NWP doubling the resolution requires approximtely 8 times more computation~\citep{bauer2015quiet}. An importance difference here is that the performance of NWP is inherently tied to the spatial resolution of the model since increased resolution allows more physical phenomenon to be directly resolved. In MetNet on the other hand the performance is not directly tied to the resolution since the network can learn to represent any sub-resolution structure in the hidden layers in a distributed way.

\section{Experiments}

We design three sets of experiments to evaluate the performance and characteristics of MetNet's predictions.  

\subsection{Eight Hour Forecasts}
\label{sec:results:methodcomparsion}
In the first set of experiments, we evaluate the performance of MetNet on the precipitation rate forecasting benchmark using the data collected in \cite{agrawalmachine} that cover the continental US. We compare MetNet with NOAA's current HRRR system, with a strong optical flow method\cite{ayzel2019optical} and with a persistence baseline using the F1 score on three precipitation rate thresholds: 0.2 mm/h, 1 mm/h and 2 mm/h in Figure~\ref{fig:F1results} (see Appendix \ref{app:hrrr_comparison} and \ref{app:opticalflow_comparison} for details).  HRRR generates forecasts covering the same region as MetNet once an hour for up to 18 hours into the future at a native  resolution of 3 km$^2$. Since MetNet outputs probabilities, for each threshold we sum the probabilities along the relevant range and calibrate the corresponding F1 score on a separate validation set.
MetNet outperforms HRRR substantially on the three  thresholds up to a lead time of respectively 400, 440 and the full 480 minutes. MetNet is also substantially better than the optical flow method and than the persistence baseline for all lead times. The F1 score degrades for higher precipitation rate thresholds for all methods since these events become increasingly rare.
Recent work using neural networks for precipitation forecasting focuses on lead times between 60 and 90 minutes \citep{agrawalmachine,ayzel2019all,lebedev2019precipitation, xingjian2015convolutional, shi2017deep} with optical flow at times outperforming neural networks ~\citep{ayzel2019all,lebedev2019precipitation}.
To our knowledge MetNet is the first machine learning model to outperform HRRR and optical flow methods on a richly structured weather benchmark at such a scale and range.

\subsection{MetNet Ablation Experiments}
\label{sec:results:ablation}

Next we perform ablation experiments to shed light on the importance of capturing spatial and temporal context and the importance of the various data sources in the input.
The first ablation experiment reduces the spatial size of the input patch to 512 km. The very first convolutional layer in the spatial downsampling part of MetNet is removed and all else is kept exactly the same.
The performance of this configuration, called MetNet-ReducedSpatial, is similar to MetNet up to 150 minutes and then it starts to become progressively worse. This indicates the importance of the large spatial context used as input as well as the ability of MetNet's architecture to capture information contained in the original receptive field of 1024 km. This contrasts with other neural networks used for 1 hour precipitation forecasting that have a U-Net-style architecture \citep{ronneberger2015u,ayzel2019all, agrawalmachine}. The receptive field of these networks at the border of the target patch is limited and likely hurts their performance and suitability for the task.

The second ablation configuration is called MetNet-ReducedTemporal and reduces the temporal context of MetNet's input features from 90 minutes prior to $T_{\bm{x}}$ to 30 minutes  prior to $T_{\bm{x}}$. This does not affect MetNet's performance significantly and  suffices to capture the advection in the input patch.
In the MetNet-GOESOnly configuration, we evaluate the contribution of the MRMS data and the ability of MetNet to predict precipitation rate from just the globally available GOES-16 data. Despite starting off substantially worse, MetNet-GOESOnly's performance approaches that of the full MetNet configuration with increasing hours of lead time suggesting that MRMS data becomes less necessary with time.

\subsection{Visualization}
\label{sec:results:qualitative}

In this section we visualize the 1 mm/h precipitation rate predictions of MetNet\footnote{Full videos: \url{https://tinyurl.com/metnet-videos}}. Table \ref{sec:results:visuals} includes predictions every two hours for up to 8 hours for MetNet, HRRR and MRMS. MetNet's predictions demonstrate substantial precipitation rate increases and decreases across time in a given location. In particular, MetNet is able to capture the creation of a significant region of higher precipitation as seen towards the middle of the US map for sample 1. With increasing lead time MetNet's predictions become increasingly blurry explained by the increased uncertainty over exact time and location of precipitation events. 

\begin{table}[]
\newcommand{\centered}[1]{\begin{tabular}{l} #1 \end{tabular}}
\newcommand{\comparerow}[2]{\centered{#1} & \centered{\includegraphics[scale=.3]{figures/samples/mrms_#2.png}} & \centered{\includegraphics[scale=.3]{figures/samples/metnet_#2.png}} & \centered{\includegraphics[scale=.3]{figures/samples/hrrr_#2.png}}}
\renewcommand{\tabcolsep}{0pt}
\begin{tabular}{rccc}
     & \textbf{Ground Truth} & \textbf{MetNet} & \textbf{HRRR} \\
     &                       & \textbf{Sample 1} &  \\
\comparerow{2 hr}{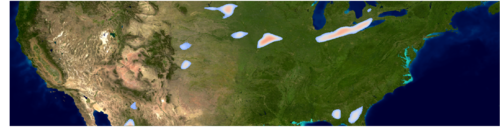} \\
\comparerow{4 hr}{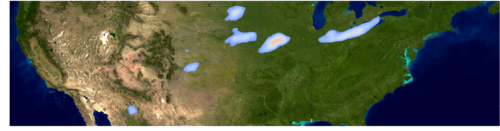} \\
\comparerow{6 hr}{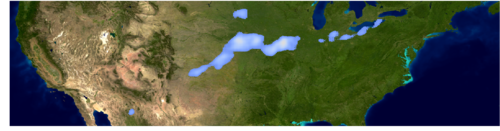} \\
\comparerow{8 hr}{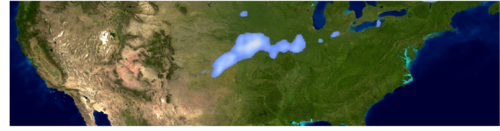} \\
     &                       & \textbf{Sample 2} &  \\
\comparerow{2 hr}{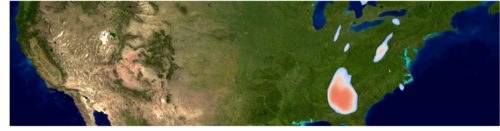} \\
\comparerow{4 hr}{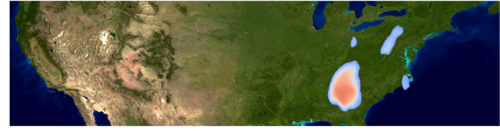} \\
\comparerow{6 hr}{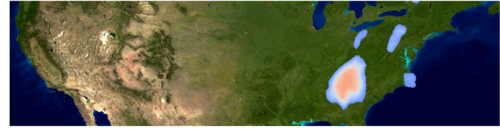} \\
\comparerow{8 hr}{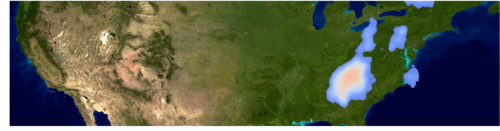} \\
\multicolumn{4}{c}{\hspace{1.cm}\includegraphics[scale=.5]{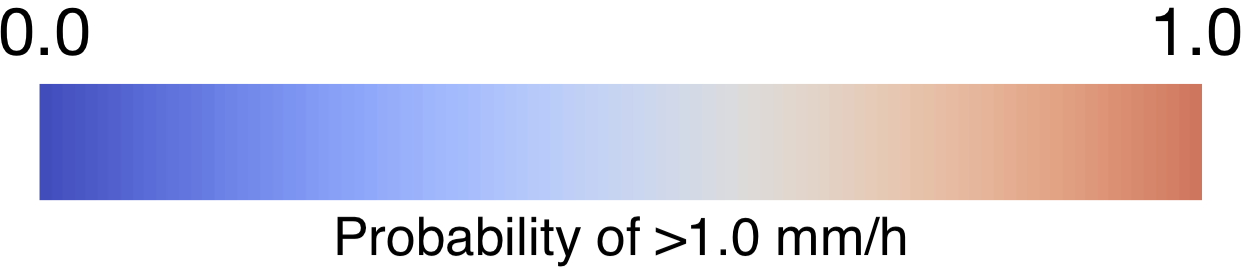}} \\
\multicolumn{4}{c}{\hspace{1.4cm} \textbf{Sample 1} \hspace{5.cm}\textbf{Sample 2}} \\
\multicolumn{4}{c}{\hspace{0.5cm}
    \begin{subfigure}[b]{.43\textwidth}
        \includegraphics[width=1\textwidth]{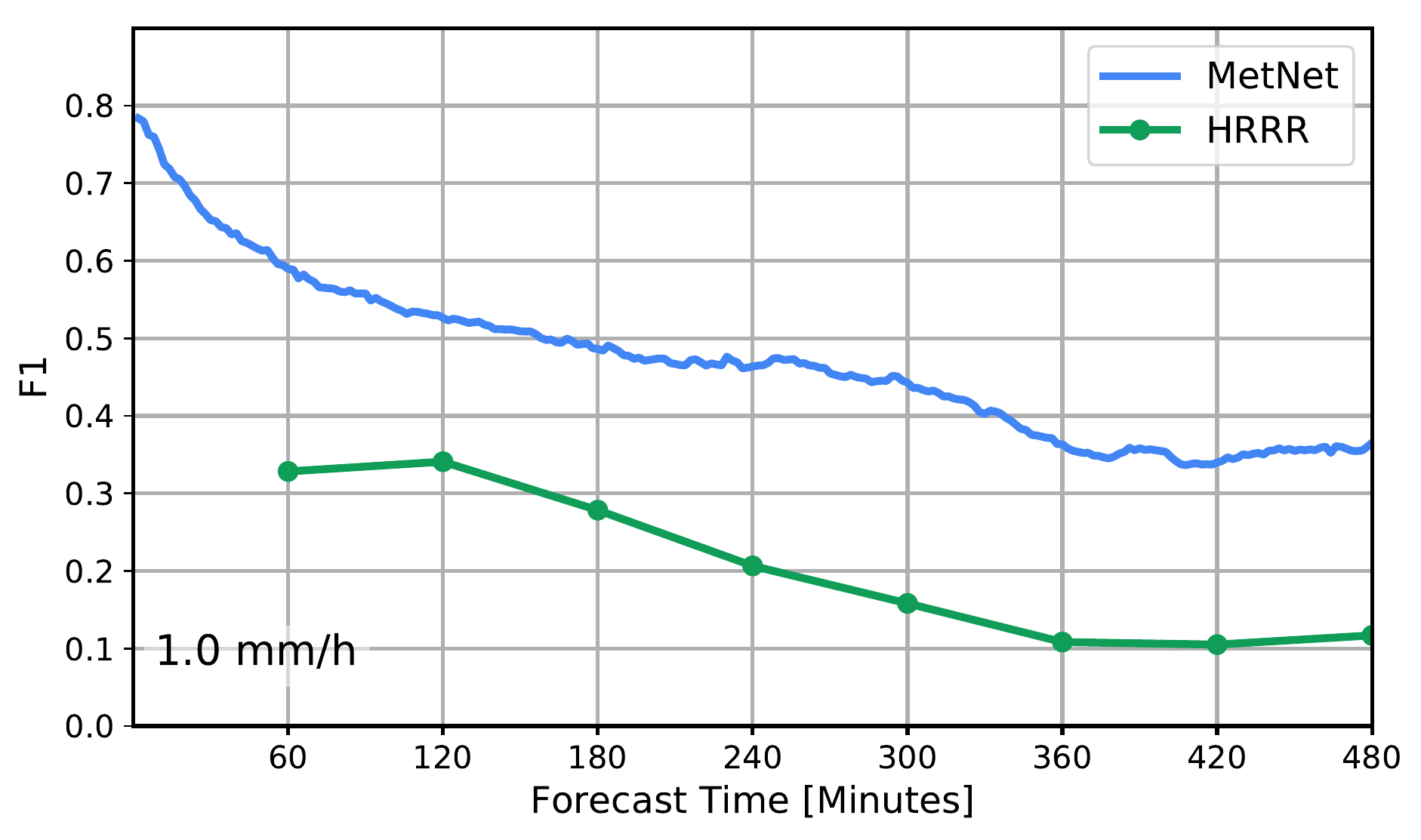}%
    \end{subfigure}
    \begin{subfigure}[b]{.43\textwidth}
        \includegraphics[width=1\textwidth]{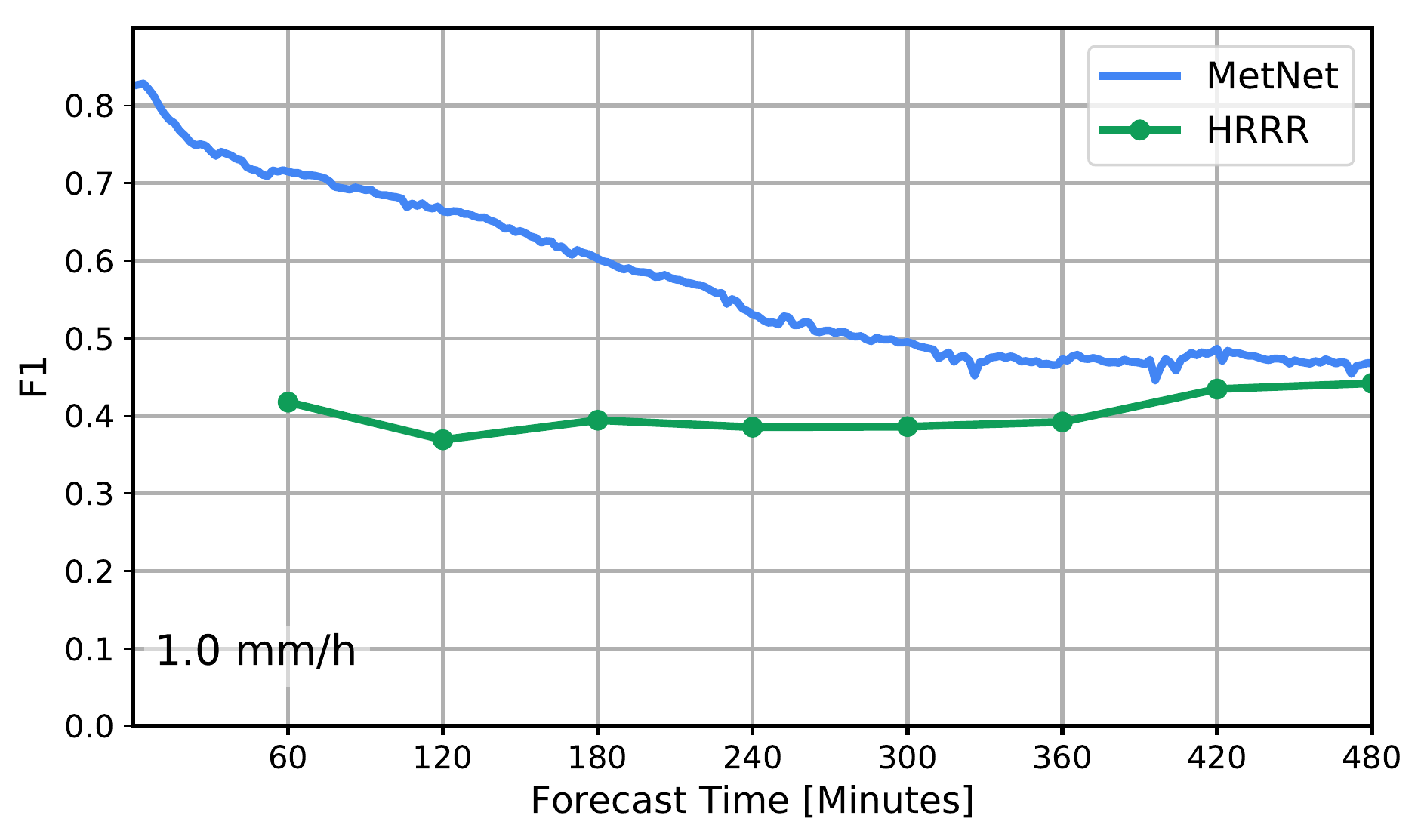}%
    \end{subfigure}%
} \\
\end{tabular}
\caption{\label{sec:results:visuals}Visualization of ground truth MRMS and predictions from MetNet and HRRR at a 1 mm/h precipitation rate threshold. For MRMS and HRRR we show pixels above the threshold whereas for MetNet we show the probability of each pixels being above the threshold.}
\end{table}

\section{Conclusion}
In this paper we presented MetNet, a neural weather model for precipitation forecasting. MetNet improves upon the current operational NWP system HRRR for up to 8 hours of lead time. Reaching beyond 8 hours will require ever larger input contexts, rigorous engineering and deeper neural networks.  

\section*{Acknowledgements}
We would like to thank Manoj Kumar, Wendy Shang, Stephen Hoyer, Lak Lakshmanan, Rob Carver, Aaron Bell, John Burge and Cenk Gazen for comments on the manuscript and insightful discussions.

\vskip 0.2in
\bibliography{references}
\appendix
\clearpage
\section{Data}
\label{app:data}

\begin{figure}[h]
    \centering
    \includegraphics[width=0.5\textwidth]{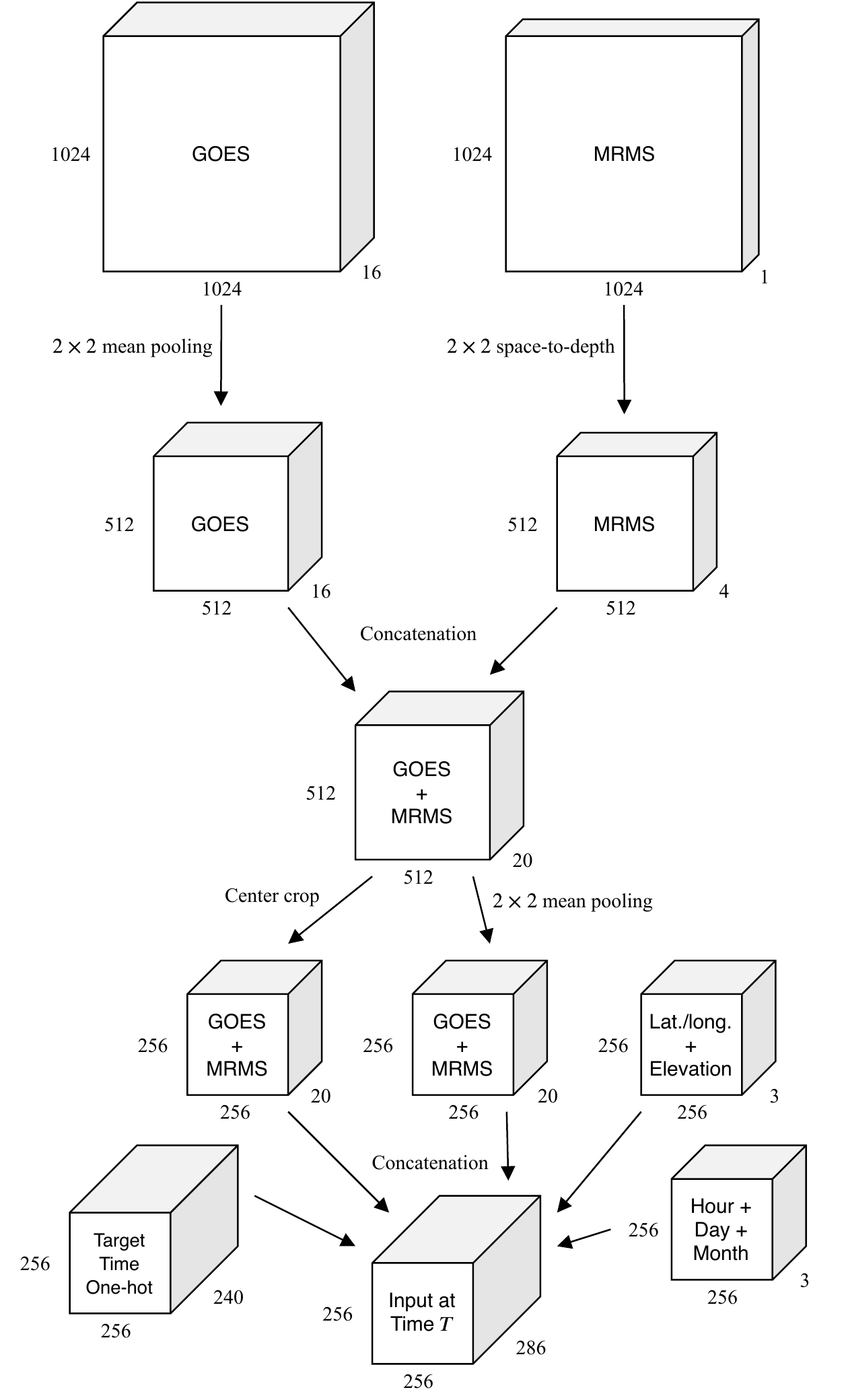}
    \caption{\label{fig:downsampling}MetNet input data downsampling and space-to-depth transformation details as well as the resulting dimensions of the input features.}
\end{figure}

\begin{figure}[h]
\includegraphics[width=0.7\textwidth]{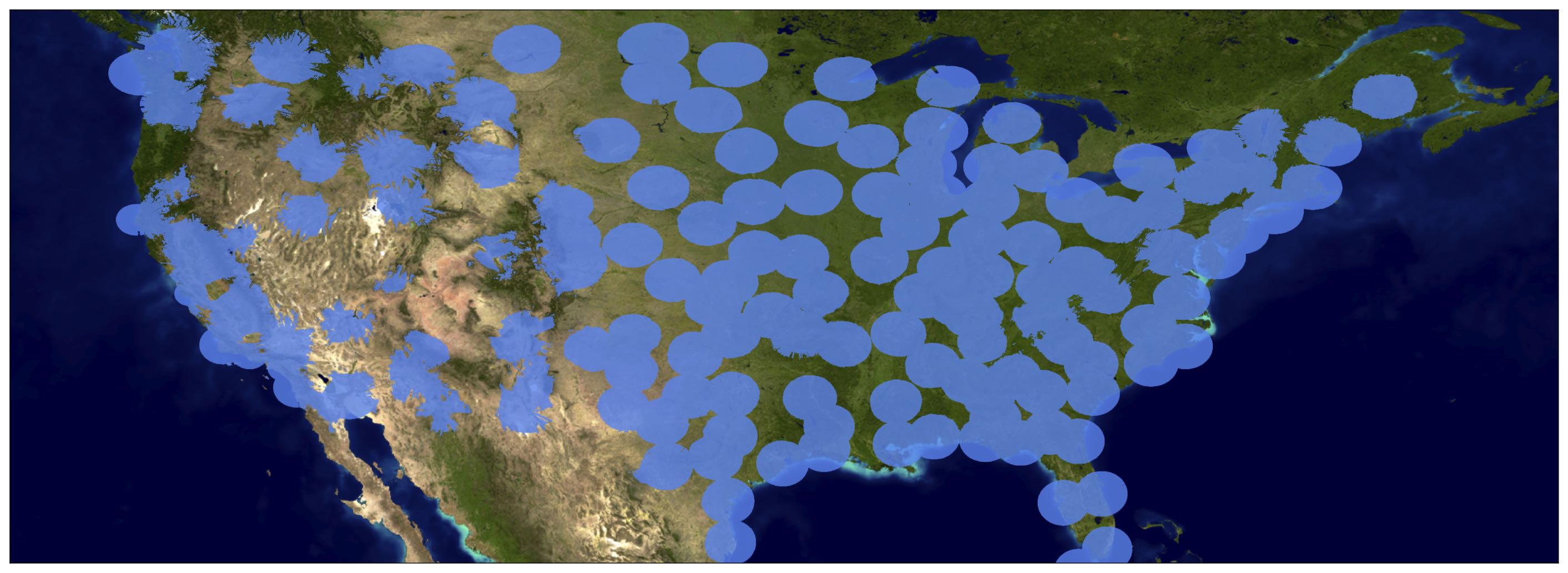}
\caption{\label{fig:mask}
    NOAA better quality masks. The loss is only calculated for targets covered by the mask to minimize issues with wrongly labelled tagets.}
\end{figure}

We construct the input and target data by selecting data relative to an anchor time $T_x$ where we stop collecting input data. We define $T_x$ as the acquisition times of the optical satellite GOES data which is sampled every 10 to 15 minutes. For the input data we find the first MRMS and GOES immediately prior to times $[-90\,\mathrm{min}, -75\,\mathrm{min} ,...., 0\,\mathrm{min}]$ relative to $T_x$. We ensure that the distributions of the input data are approximately normal distributed and within the a $[-1, 1]$ range by applying the following data normalization: The GOES data is robustly normalized by subtracting the median and dividing by the inter-quantile range and the MRMS data is transformed by $\log(x + 0.01)/4$. To avoid any remaining outliers we replace all NaNs with zeros and squash the data to $[-1, 1]$ using the hyperbolic tangent function. As additional static features we include longitude, latitude, elevation and time (represented as three normalized features: $\mathrm{hour}/24$, $\mathrm{day}/31$, $\mathrm{month}/12$) as feature maps. To reduce the memory footprint on the model we perform space-to-depth transformation or downsampling of the input, see Figure~\ref{fig:downsampling} for details. The prediction targets are the MRMS measurements immediately after times  $[2\,\mathrm{min}, 4\,\mathrm{min} ,...., 480\,\mathrm{min}]$ relative to $T_x$. We only retain samples where all GOES and MRMS could be mapped to the desired sampling times with a maximum of 5 minutes mismatch. Furthermore the range of each radar in the MRMS data source is approximately $250\,\mathrm{km}$ for an unblocked field of view with data quality generally dropping with distance to the radar. To minimize issues with wrongly labelled targets we only assign a loss to pixels contained in the NOAA \textit{better} quality map (Figure~\ref{fig:mask}). All data are remapped to the NOAA \textit{CONUS} rectangular grid of size $7000 \times 2500$ approximately covering the continental US from $-135^{\circ}$ to $-65^{\circ}$ longitude and $25^{\circ}$ to $50^{\circ}$ latitude with a $0.01^{\circ}$ degree resolution. 
For both MRMS and GOES we acquired data for the period January 2018 through July 2019. We split the data temporally into three non-overlapping data sets by repeatedly using approximately 16 days for training followed by two days for validation and two days for testing. From these temporal splits we randomly extracted 13,717 test and validation samples and kept increasing the training set size until we observed no over-fitting at 1.72 million training samples. For ablation studies we leave the target unchanged and only crop the spatial or temporal extend of the input data. All models and data-processing were implemented in a combination of TensorFlow~\citep{tensorflow2015-whitepaper} and JAX~\citep{jax2018github} using up to 256 Google TPU accelerators in parallel.

\FloatBarrier

\FloatBarrier

\section{Numerical Weather Prediction}
\emph{Numerical Weather Prediction} is the most successful framework to perform medium- and long-range (up to 6 days with high confidence) forecast to date~\citep{bauer2015quiet}.  The core of NWP models are a set of PDEs and other equations which condenses our current beliefs about the dynamical behaviors in the earths atmosphere~\citep{wicker2013everything}. The exact formulation of NWP models varies depending on the use case, target domain and scale, but the model variables includes surface-based, airborne and satelite-based measurements, such as temperature, moisture, precipitation, wind fields and pressure~\citep{ecmwf2018statement}. Besides the validity of the mathematical formulation, the accuracy relies on the quality of the initial conditions---that is, how faithfully the initial conditions resemble the current state of the atmosphere. The initial conditions in modern NWP models are estimated through a complex \textit{data assimilation} process leveraging a wide range of available information, including past and asynoptic observations, to improve initialization~\citep{barker2004three}. Despite such effort, initial condition errors do always exist~\citep{NAP6434} and negatively impact model performance during the \textit{spin-up period}~\citep{hwang2015improved}, leading to suboptimal short-range forecast. In addition solving the PDEs numerically is computationally demanding with runtimes of more than one our for regional models such as HRRR. 

\section{HRRR Baseline}
\label{app:hrrr_comparison}
We primarily compare against the current operational NOAA HRRR version 3~\citep{benjamin2016north} NWP model. HRRR produces forecasts each hour for 1-18 hours into the future at a 3-by-3 kilometer resolution. The HRRR predictions are resampled to the NOAA CONUS grid described in Appendix~\ref{app:data}. We compare the HRRR precipitation rate forecasts with the nearest MRMS measurement discarding samples where a MRMS target could not be found within 5 minutes of the HRRR lead time. 

\section{Optical Flow Baseline}
\label{app:opticalflow_comparison}
\begin{figure*}
    \begin{subfigure}[b]{.49\textwidth}
        \centering
        
        \includegraphics[width=1\textwidth]{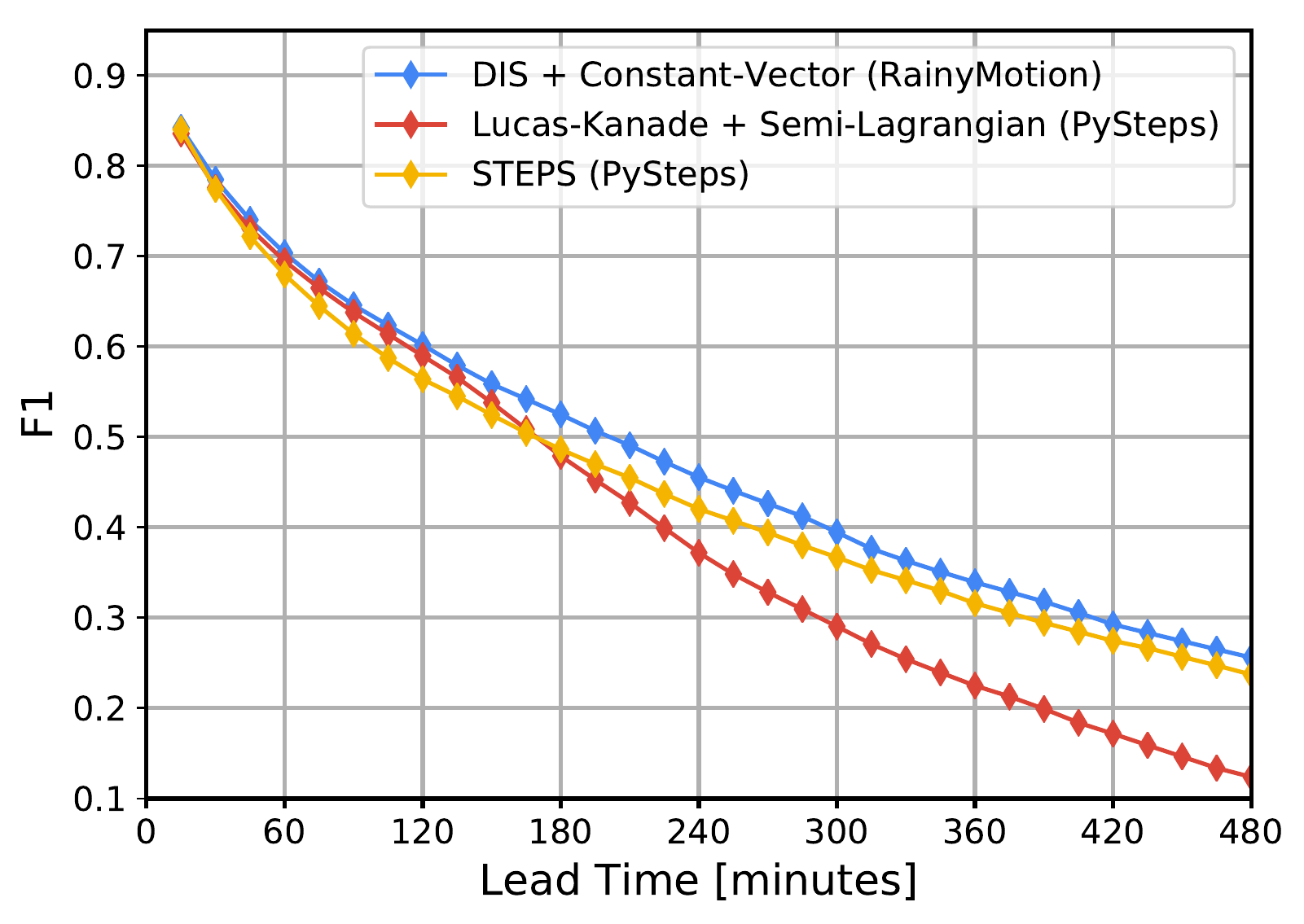}
        
    \end{subfigure}\,\,\,%
    \begin{subfigure}[b]{.49\textwidth}
        \centering
        \includegraphics[width=1\textwidth]{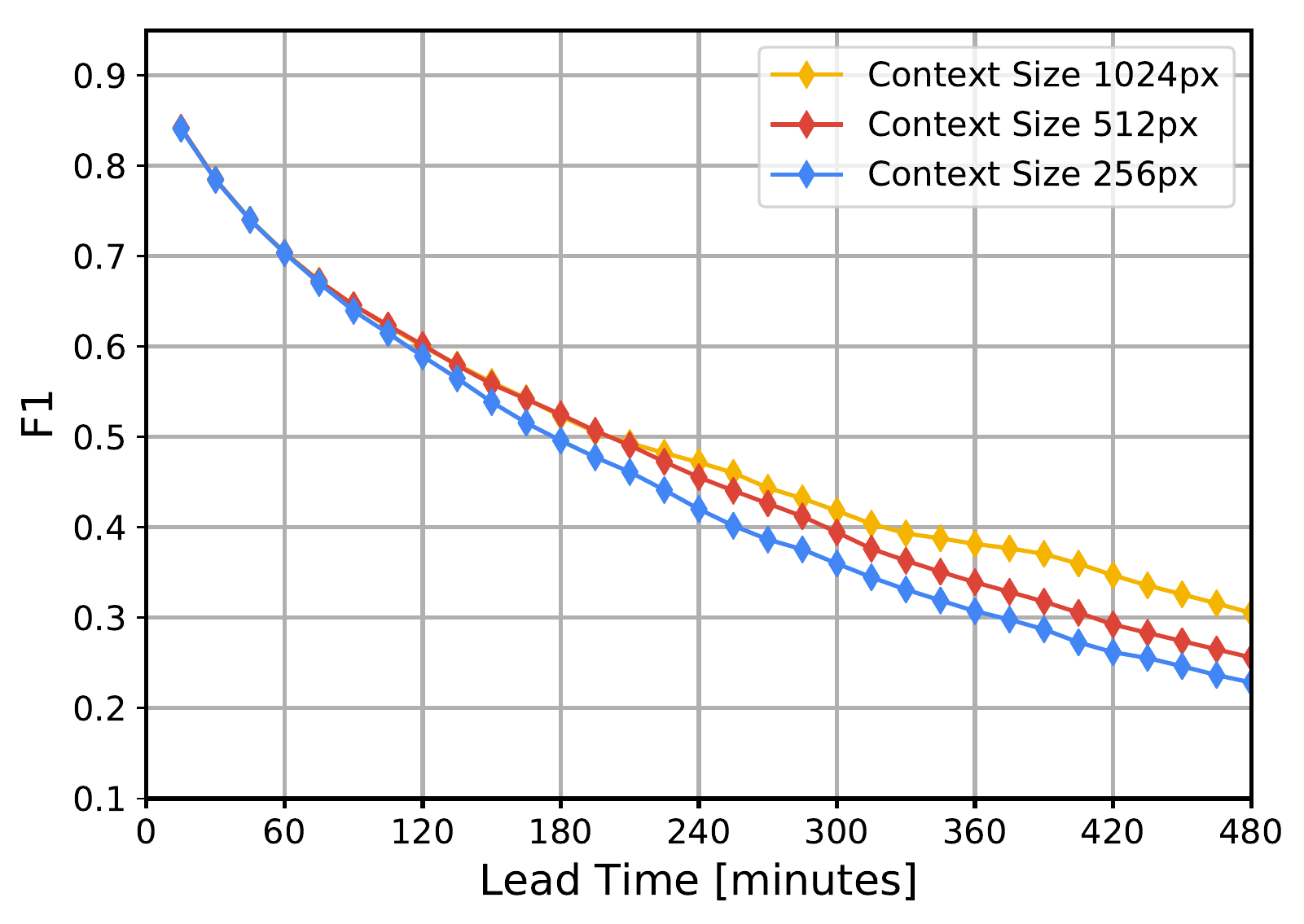}%
    \end{subfigure}%
    \caption{\label{fig:opticalflow_comparison}Comparison of optical flow baselines and effect of context size using $0.1$ mm/h as cutoff level. \textbf{Left:}~Comparison of Dense Inverse Search using constant-vector scheme for extrapolation (DIS~+~Constant-Vector), Lucas-Kanade with semi-Lagrangian scheme for extrapolation (Lucas-Kanada~+~Semi-Lagrangian) and the Short‐Term Ensemble Prediction System (STEPS) using 512 pixels of spatial context. Implementation libraries are given in parenthesis in the legend. \textbf{Right:}~The Dense Inverse Search based algorithm with different context sizes demonstrating that adding more spatial context always improves performance.}
\end{figure*}

To establish a strong optical flow baseline model we benchmarked several algorithms for precipitation rate forecasting available in the RainyMotion~\citep{ayzel2019optical} and PySteps~\citep{pulkkinen2019pysteps} libraries. For all methods we only use the MRMS data to fit the tracking algorithm, but otherwise leave the data unchanged. In Figure~\ref{fig:opticalflow_comparison}~(left) we compare three optical flow based methods for precipitation rate forecasting. We found that using a dense inverse search algorithm~\citep{kroeger2016fast} for tracking and a backward constant-vector scheme for extrapolation~\cite{bowler2004development} performed better than both the classical Lucas-Kanade~\citep{lucas1981iterative} tracking coupled with a backward semi-Lagrangian extrapolation scheme or the probabilistic Short‐Term Ensemble Prediction System~\citep{bowler2006steps, seed2013formulation}.  Figure~\ref{fig:opticalflow_comparison}~(right) shows the effect of reducing the context size for optical flow. Similar to the findings in Figure~\ref{fig:ablations} adding more spatial context is always beneficial especially for the longest lead times. In summary we found the algorithm using Dense Inverse Search with constant-vector scheme extrapolation and a spatial context size of $1024\times1024$ to perform the best and use that for all optical flow results in the main paper.

\end{document}